\documentclass[
]{ceurart}

\usepackage{listings}
\begin{document}

\copyrightyear{2024}
\copyrightclause{Copyright for this paper by its authors.
  Use permitted under Creative Commons License Attribution 4.0
  International (CC BY 4.0).}

\conference{In: R. Campos, A. Jorge, A. Jatowt, S. Bhatia, M. Litvak (eds.): Proceedings of the Text2Story'24 Workshop, Glasgow (Scotland), 24-March-2024}

\newcommand{\red}[1]{\textcolor{red}{#1}}

\title{Evaluating the Ability of Computationally Extracted Narrative Maps to Encode Media Framing}

\author[1]{Sebastian {Concha Macías}}[%
email=sebastian.concha@alumnos.ucn.cl
]
\address[1]{Universidad Católica del Norte,
  Av. Angamos 0610, Antofagasta, 1270709, Chile.}

\author[1]{Brian {Keith Norambuena}}[%
orcid=0000-0001-5734-8962,
email=brian.keith@ucn.cl,
url=https://briankeithn.github.io/,
]
\cormark[1]

\cortext[1]{Corresponding author.}

\begin{abstract}
Narratives serve as fundamental frameworks in our understanding of the world and play a crucial role in collaborative sensemaking, providing a versatile foundation for sensemaking. Framing is a subtle yet potent mechanism that influences public perception through specific word choices, shaping interpretations of reported news events. Despite the recognized importance of narratives and framing, a significant gap exists in the literature with regard to the explicit consideration of framing within the context of computational extraction and representation. This article explores the capabilities of a specific narrative extraction and representation approach---narrative maps---to capture framing information from news data. The research addresses two key questions: (1) Does the narrative extraction method capture the framing distribution of the data set? (2) Does it produce a representation with consistent framing? Our results indicate that while the algorithm captures framing distributions, achieving consistent framing across various starting and ending events poses challenges. Our results highlight the potential of narrative maps to provide users with insights into the intricate framing dynamics within news narratives. However, we note that directly leveraging framing information in the computational narrative extraction process remains an open challenge.
\end{abstract}


\maketitle

\section{Introduction}
Narratives provide a fundamental framework to our understanding of the world \cite{abbott2008cambridge} and are integral to human relations \cite{miskimmon2014strategic}. In particular, narratives serve as cognitive tools enabling individuals to forge connections between seemingly disparate events, playing a crucial role in collaborative sensemaking within society \cite{keith2020maps}. Narratives also provide the foundation for versatile sensemaking tools in various domains, including news and intelligence analysis \cite{stasko2008sensemaking}. In this context, constructing meaningful representations from sources such as news data or intelligence reports facilitates a deeper understanding of the intricate storylines and the overarching narrative contained in data \cite{keith2023survey}.

In conjunction with narratives, framing emerges as a potent mechanism for shaping public opinion and perception. In general, framing refers to the process by which information is selectively presented to influence the perception and opinion of the audience. It involves focusing on certain aspects of a topic through the use of particular words, images, or approaches, to highlight certain aspects and downplay others \cite{entman1993framing}. 

Throughout this work, we focus on \textit{media framing} in textual news data \cite{scheufele1999framing}, as opposed to other types of framing used in the literature and non-written media. Operating subtly through specific word choices, media framing influences readers by highlighting particular facets of an event, thereby molding the audience's interpretation of reported news \cite{hamborg2019illegal}. For example, the choice of terms, such as ``illegal alien'' versus ``undocumented immigrant'', shows this influence in discussions around immigration. Various approaches to scrutinizing media framing have been developed by communication scholars, with a common quantitative method involving the previous definition of frames and subsequent manual classification of news articles into identified frames through machine learning techniques \cite{liu2019detecting}. 

A notable gap in the literature involves the absence of studies explicitly addressing framing within the domain of computational extraction and representation of narratives from news data \cite{keith2023survey}. While numerous research works explore narrative construction and representation in the news domain, a critical aspect often overlooked is the explicit consideration of media framing elements. The interplay between computational methods and the nuanced influence of framing on information interpretation remains largely unexplored.

In this context, the purpose of this work is to evaluate whether a specific computational narrative representation---narrative maps \cite{keith2020maps}---and the associated extraction method can properly capture framing information from news data. Thus, our research asks the following research questions:
\begin{itemize}
    \item \textbf{RQ1:} Does the narrative maps extraction algorithm properly capture the framing distribution of the data set?
    \item \textbf{RQ2:} Does the narrative maps extraction algorithm produce a representation with consistent framing? 
\end{itemize}
We answer these questions by testing the extraction algorithm on a news data set with framing labels. In particular, we explore how the extraction algorithm behaves under different combinations of framing for the starting and ending events, which act as seeds for the extraction method. Our results suggest that the extraction method is capable of capturing the same framing distribution of the data set, but that it does not necessarily produce a representation with consistent framing when considering the starting and ending events.

\section{Background and Related Work}
\subsection{Narrative Maps and Extraction}
Narrative maps are a visual representation of storylines and events based on a cartographic metaphor to model the information landscape \cite{keith2020maps, shahaf2015information}. These maps are designed to illustrate the relationship between different events, characters, themes, and key points that make up a story or a series of events in a graphical format. Computationally, narrative maps are represented as directed acyclic graphs, where each node represents an event.

In the context of narrative analysis applied to computational journalism or other fields, narrative maps serve as tools to break down the narrative structure of news or events \cite{keith2020maps}. They use graphical elements, such as nodes and connections, to show how the story unfolds, how events are interconnected, and how they contribute to the overall understanding of the narrative.

These maps are represented as directed acyclic graphs \cite{keith2020maps}, where nodes represent key events or moments, and the connections between them depict the sequence or relationship between those events. Additionally, they may incorporate interactive features or layers of artificial intelligence to allow the model to adapt to user needs and provide explanations about the narrative structure \cite{keith2023iui}.

\subsection{Computational Frame Modeling}
Frame extraction methods tend to focus on \textit{semantic frames}, where extensive research has led to the development of methods such as FrameNet \cite{baker1998berkeley} and extraction tools like Open-SESAME \cite{swayamdipta2017frame}, which use deep learning models alongside syntactic information to extract semantic frames from the text. However, these approaches are too granular in scope, as semantic frames focus more on linguistic aspects, rather than applicable concepts in the news domain. Thus, researchers have proposed the concept of \textit{media frames} \cite{card2015media}.

Researchers have used various kinds of models to analyze media frames \cite{naderi2017classifying, hamborg2019automated}. Some authors follow an unsupervised approach, where frames are identified from an unlabeled set of data. Others have carefully constructed labeled data sets to train machine learning classifiers. Another aspect of this task that must be considered is the level of granularity of the framing analysis. Some works focus on article-wide high-level frames \cite{liu2019detecting, guo2023proposing}, while others explore sentence-level framing, or even both \cite{naderi2017classifying}.

However, as mentioned before, despite the significant progress made in developing methods for narrative construction and representation in the news domain, the explicit consideration of media framing elements in the context of computational narrative extraction and representation from news data remains largely unexplored \cite{keith2023survey}.

For the purposes of this article, we use a pre-labeled data set \cite{liu2019detecting} to check whether the narrative extraction algorithm is capable of capturing framing information. We do not attempt to extract frames automatically at this stage, as the main goal is to first identify the capabilities of the baseline model. 

\section{Materials and Methods}
\subsection{Data Set Description}
We use the Gun Violence Frame Corpus (GVFC) from Liu et al. \cite{liu2019detecting}. The data set consists of 1300 news articles in English from multiple U.S. based sources extracted during the year 2018. The GVFC focuses on media frames commonly used when reporting the issue of Gun Violence. In particular, it has 9 types of frames (including both issue-specific and generic frames) \cite{liu2019detecting}. Table \ref{dist} shows the distribution of frame labels in the data.

\begin{table}[!htbp]
\centering
\footnotesize
\caption{Distribution of each frame label (main media frame) in the Gun Violence Frame Corpus \cite{liu2019detecting}.}
\begin{tabular}{ccl}
\toprule
\textbf{ID: Frame Label [Reference]}             & \textbf{Freq.}  &\textbf{Example Headline}\\ \midrule
1: 2nd Amendment \cite{birkland2009media}              & 38                           &''Breaking down the 27 words of the Second Amendment``\\  
2: Gun Control/Regulation \cite{birkland2009media}     & 215                          &''How gun background checks work``\\  
3: Politics \cite{schnell2001assessing}               & 373                          &''Romney open to new gun measures``\\  
4: Mental Health \cite{defoster2018guns}             & 65                           &''Florida shooter a troubled loner with white supremacist ties``\\  
5: School/Public Space Safety \cite{haider2001gun}& 137                          &7 ways to help prevent school shootings\\  
6: Race/Ethnicity \cite{leavy2009american}            & 114                          &''Lawyers call US gun charges for Mexican man 'vindictive'``
\\  
7: Public Opinion \cite{neuman1992common}            & 237                          &``Students call for action after Florida school mass shooting''\\  
8: Society/Culture \cite{lawrence2004guns}           & 41                           &``White House weighs video game link to gun violence''\\ 
9: Economic Consequences \cite{neuman1992common}     & 80                           &``Calstrs to engage with assault-weapon sellers first, divest last''\\ \bottomrule \
\end{tabular}
\label{dist}
\end{table}
We note that the narrative extraction method has a high computational cost as the data set size increases. Thus, we decided to decrease the data set size from 1300 articles while maintaining the original distribution of the frame labels. Since this data set is unbalanced with respect to the frame labels, we performed stratified sampling \cite{parsons2014stratified} and obtained a new data set of 131 articles. 

Furthermore, we are faced with the absence of essential variables for the narrative extraction method. Specifically, the algorithm requires temporal information, but the original data set does not contain any timestamps to guide the extraction process. Furthermore, the data set did not provide the source of the news, which is also required by the extraction algorithm to generate the final visualization. Thus, we had to manually search for the original sources of each article based on its headline and add the missing information to the data set.

Moreover, before extracting the narrative representation from our new subset of the original data set, we decided to reduce the complexity of the framing model to account for our smaller data set size. Thus, we reduced the number of frames into three higher-level frames by grouping related frames into a single abstract category. This simplification is necessary to ensure that the framing model remains effective and accurate, despite the reduced data size. By grouping related frames into broader, more abstract categories, we maintain the essence of the original framing structure while making it more manageable and suitable for the limited data set.

In particular, for the first, second and third frames of the data set, we note that their main focus is on political issues, mostly related to gun control. Therefore, we united them under the name ``Frame 1: Political Issues''. Likewise, the fourth and fifth frames focused on aspects related to mental healthcare issues, as well as school and public safety. Both of these elements could be addressed by public services. Hence, we united them under the name ``Frame 2: Public Services''. Finally, the last four frames were oriented towards either cultural issues or societal issues, including discussions around race and ethnicity, public opinion, and economic consequences. Thus, we united them under the name ``Frame 3: Cultural and Societal Issues''. Finally, this led us to simplify the set of nine possible frames to just three frames \footnote{The data set is available here: \url{https://doi.org/10.5281/zenodo.10822952}}.

\subsection{Narrative Extraction Method}
\begin{figure}
    \centering
    \includegraphics[width=0.7\textwidth]{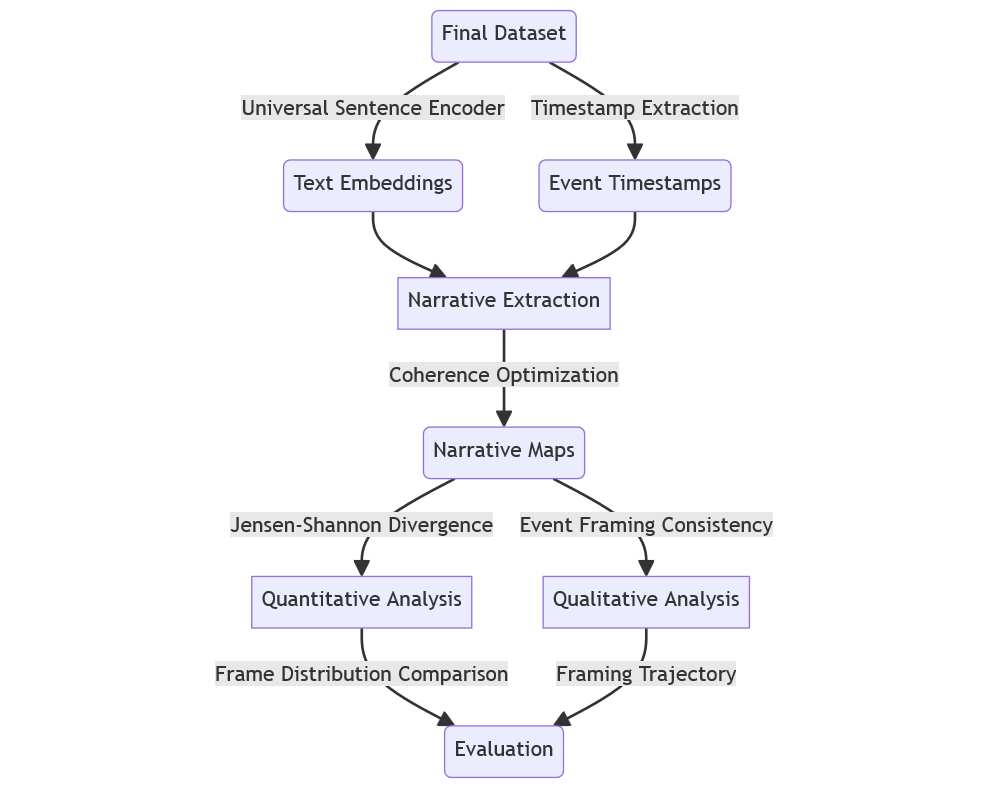}
    \caption{Technical methodological diagram illustrating the key steps in the narrative extraction and analysis process, starting with the preprocessed final data set and ending with the quantitative and qualitative evaluation of the resulting narrative maps.
}
    \label{fig:process}
\end{figure}

In this study, we employ a narrative extraction method developed by Keith and Mitra \cite{keith2020maps} to generate narrative maps from textual news articles. A narrative map is a computational representation of a narrative in the form of a directed acyclic graph, where nodes represent events and edges represent the chronological order and causal relationships between events. Figure \ref{fig:process} shows a representation of the general extraction and subsequent analysis process.

From a computational standpoint, the extraction method takes a list of news articles, where each news article is represented as a pair of \textit{text embeddings} and the publication date as a  \textit{timestamp}. Following an optimization procedure, the method generates a narrative map that captures the key events and their connections within the narrative. For the purposes of our experiment, we only use the headline to represent the news article, as it contains the most relevant information for framing purposes, under the assumption that breaking news articles tend to follow the inverted pyramid structure \cite{norambuenaevaluating}. For the embeddings, we use the Universal Sentence Encoder v4.0 from TensorFlow \cite{cer2018universal}, which is appropriate for short texts such as news headlines, based on the same implementation of the original extraction method \cite{keith2020maps}.

\begin{figure}[htbp]
    \centering
    \includegraphics[width=0.66\textwidth]{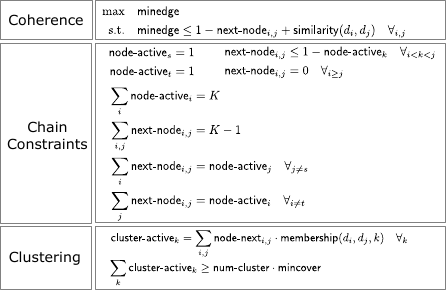}
    \caption{Linear programming formulation of the extraction method of Keith and Mitra \cite{keith2020maps}.}
    \label{fig:lp}
\end{figure}

Regarding the extraction process itself, this method leverages an optimization approach based in maximizing coherence---assessing the logical flow and ensuring that the storylines make sense. In particular, we impose structural and topic coverage constraints through linear programming, adopting the methodology outlined by Keith and Mitra \cite{keith2020maps}. The formulation of the linear program governing the extraction method is presented in Figure \ref{fig:lp}.

Our interpretation of coherence is based on the concept of similarity, grounded in the rationale that connected events should exhibit minimal fluctuations in topics or content throughout the narrative. More precisely, the coherence value, indicating the logical continuity of joining two events, is computed by evaluating both text similarity---based on an embedding representation---and topical similarity---derived from topical clusters computed in a 2D projection space generated by the UMAP algorithm \cite{mcinnes2018umap} with the HDBSCAN method \cite{mcinnes2017hdbscan}.

The structural constraints shown in Figure \ref{fig:lp} play a crucial role in ensuring the generation of a directed acyclic graph featuring a singular source and a single sink interconnected in chronological order through multiple storylines. Furthermore, the extraction algorithm relies on a user-defined parameter for the expected length of the main storyline ($K$). This parameter also directly influences the overall size of the map. Throughout our experiments, we used a parameter $K = 6$, as it provides maps of appropriate complexity for our data set size. This value is based on empirical observations from previous studies and our own preliminary experiments, striking a balance between capturing key events and maintaining a manageable narrative size. 

The topic coverage constraints ensure that the extracted narrative encompasses a minimum percentage of the topics present in the data, determined by a predefined coverage threshold. The topic coverage constraints are based on topic similarity information  \cite{keith2020maps}, based on the same information used to compute coherence. Throughout these experiments, we used a coverage threshold of 50\%, which allows for a sufficient level of abstraction while preserving the essential topical information of the data set. 

\subsection{Evaluation Metrics}
To quantitatively evaluate our results, we use the Jensen-Shannon divergence to measure how well the framing distribution of the extracted narrative map captures the framing distribution of the data set. In general, the Jensen–Shannon divergence allows us to measure the similarity between two probability distributions. In this case, we consider the normalized frequencies of the different frames in the data set and the maps as probability distributions to perform the comparisons. We note that this measure of similarity takes values between 0 and 1, where 0 represents similar distributions and 1 represents completely dissimilar distributions. 

\section{Results and Discussion}
\subsection{Resulting Maps}
We generated 9 maps, accounting for all the possible combinations of starting and ending frames. Table \ref{tab:map-size} shows the sizes of the maps based on the number of events contained in them. The maps have a mean size of 19.7 nodes and median size of 21 nodes. The standard deviation is 3.15 nodes. All maps have generally similar sizes in terms of covered events, but they may cover different events, based on the chosen starting and ending events. We note that despite differences in size and specific contents, the framing distribution of all these maps are quite similar, as shown in the next section.

\begin{table}[ht]
    \centering
    \caption{Map sizes of each combination starting and ending frames based on number of events in the map.}
    \begin{tabular}{cccc} \hline 
         \textbf{Map Size}&  End: Frame 1&  End: Frame 2&  End: Frame 3 \\ \hline 
         Start: Frame 1&  24 &  21&  19 \\ 
         Start: Frame 2&  16&  17&  22 \\  
         Start: Frame 3&  22&  22&  15 \\  \hline
    \end{tabular}
    \label{tab:map-size}
\end{table}

\subsection{Quantitative Results}
We show the quantitative results of our evaluation in Table \ref{tab:js-divergence}. On average, the narrative extraction algorithm captures a very similar framing distribution to the full data set. The consistently low average Jensen-Shannon divergence of 0.0218 across all tested combinations signals a notable alignment between the framing distributions of the generated narrative maps and the data set. A Jensen-Shannon divergence of 0 would indicate identical distributions, so values closer to 0 suggest high similarity. With all divergence scores below 0.1, the results support the conclusion that the narrative maps closely capture the overall framing distribution of the original data.

\begin{table}[ht]
    \centering
    \caption{Jensen-Shannon divergence values obtained by comparing the framing distribution of each combination starting and ending frames with the distribution of the data set.}
    \begin{tabular}{cccc|c} \hline 
         \textbf{J-S Divergence}&  End: Frame 1&  End: Frame 2&  End: Frame 3& \textit{Average}\\ \hline 
         Start: Frame 1&  0.0764&  0.0009&  0.0050& 0.0274\\ 
         Start: Frame 2&  0.0270&  0.0396&  0.0172& 0.0280\\  
         Start: Frame 3&  0.0069&  0.0171&  0.0063& 0.0101\\  \hline
         \textit{Average}&  0.0368&  0.0192&  0.0095& 0.0218\\ \hline
    \end{tabular}
    \label{tab:js-divergence}
\end{table}

The most dissimilar distributions occur when we start and end with \textit{Frame 1} (Map 1-1). This difference could be explained by the fact that we are selecting both the starting and ending event with the most frequent frame in the data set, which could lead to a biased narrative map that mostly focuses on that frame, guided both by the higher frequency on the whole data and the user selected events. Note that around 48.8\% of the events are assigned to \textit{Frame 1} in the data set, while 79.2\% of the events of the aforementioned map are assigned to this frame.

Nevertheless, our analysis of results suggests that the narrative maps extraction algorithm properly captures the framing distribution of the data set. Thus, based on these results, the extracted narrative maps could potentially provide users of the representation with insights regarding the framing of the news narrative by the media. 

\subsection{Qualitative Example}
\begin{figure}[htbp]
    \centering
    \includegraphics[width=0.65\textwidth]{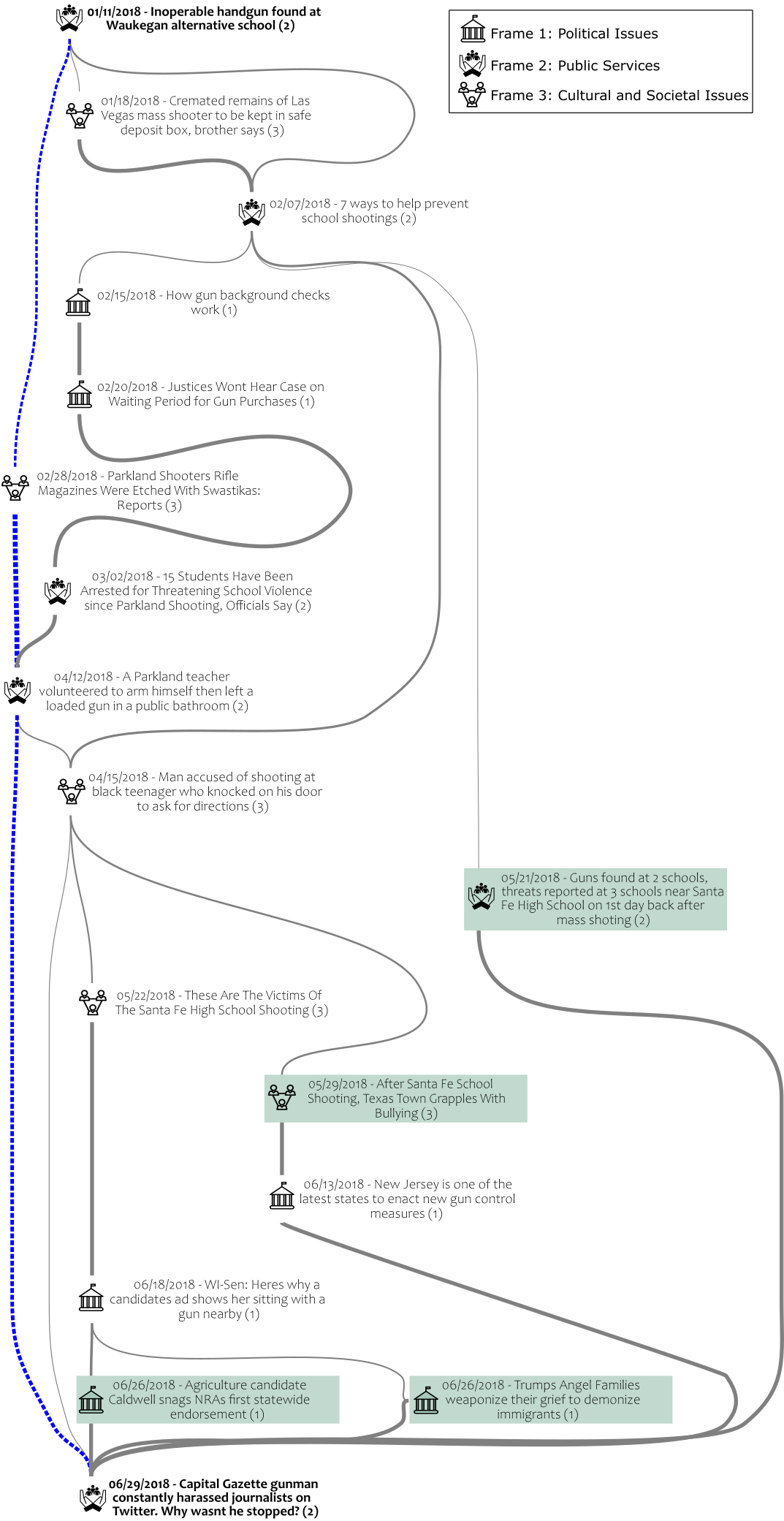}
    \caption{Sample narrative map extracted from news events spanning January to June 2018. Events are represented as nodes labeled with headlines and frames. Connections indicate narrative relationships between events. Inconsistencies in framing across the map highlight challenges in producing coherent framing narratives through the extraction process.} 
    \label{fig:FrameMap}
\end{figure}

To evaluate whether the narrative mapping algorithm produces representations with consistent framing, we examined sample narrative maps extracted from news events spanning several months. We show an example narrative map in Figure \ref{fig:FrameMap} that contains events covering school safety, political debates, and cultural impacts related to gun violence in the United States. In particular, the map had endpoints related to \textit{Frame 2} of the data set. Thus, if framing was consistent, the algorithm should select mostly documents with that particular framing. However, analysis of the narrative map shown revealed inconsistencies in how events were framed throughout the narrative.

For instance, while the initial events dealing with discovery of a handgun in a school and prevention of school shootings were framed around public service issues, later events in the map shifted focus to political debates around gun control and cultural/societal impacts of shootings. The algorithm connected the early school safety events to these later events even though they contained different frames. This suggests the narrative mapping process focused on content similarities between events, rather than maintaining consistent framing across the full map.

Additionally, we observed abrupt framing shifts between connected events at times, such as between a story on the Parkland shooter's swastika-etched magazines and one on a Parkland teacher leaving a loaded gun in a bathroom. While containing similar content, these events had different framing (cultural/societal issues vs. public services/safety). These framing inconsistencies can undermine the coherence of the extracted narrative, making it challenging to discern a clear framing trajectory across the map. Abrupt frame changes between events disrupt the logical flow and complicate interpretation of the narrative's overarching stance or perspective. The problem is not isolated to this single example; similar framing discrepancies appear across other sample maps, indicating a broader limitation of the current approach.

Based on analysis of this example and other sample narrative maps, we found the algorithm does not optimize for consistent event framing when extracting the narratives. While the technique is able to capture the framing distribution of the broader data set, maintaining coherent frames across an entire narrative remains a challenge for the approach.

\subsection{Implications}
The findings of this study have important implications for both the research community and practitioners seeking to leverage computational narrative extraction techniques to understand media framing. From a research perspective, our work highlights the need for greater attention to the role of framing in shaping the structure and coherence of extracted narratives. 

While the narrative maps approach demonstrates the ability to capture high-level framing distributions, the inconsistencies revealed in our qualitative analysis suggest that current techniques may be insufficient for generating fully coherent narrative representations. Future research should prioritize the development of novel approaches that can more effectively maintain consistent framing across narrative chains.

For practitioners, such as journalists, policy makers, or analysts, our findings underscore both the potential benefits and pitfalls of relying on computational narrative extraction to assess media framing. On one hand, the technique's ability to preserve the overall framing distribution of a large corpus of news articles points to its value as a tool for quickly identifying dominant frames and gauging their relative prominence. This high-level perspective could guide further investigation or provide a starting point for exploring framing patterns. 

However, the framing inconsistencies within individual narrative maps suggest that caution should be exercised when interpreting the specific structure and connections of these representations. Users should be aware that the narrative chain may not always reflect a coherent framing trajectory and that abrupt shifts in perspective can obscure the underlying story.

\section{Conclusions}
This work investigated the capabilities of narrative maps, a computational technique for extracting narratives from text data, to capture framing information. Framing refers to the promotion of a particular perspective in a narrative through selective wording. Our analysis focused on two key questions - whether the algorithm accurately captures framing distributions, and if it produces narratives with consistent framing. While the technique showed promise in capturing the framing distribution of the data set, we found limitations in achieving consistent framing for the extracted narratives.

In particular, our analysis of the performance of the narrative extraction algorithm shows its ability to capture framing distributions. The consistently low average Jensen-Shannon divergence of 0.0218 across all tested combinations signals a notable alignment between the framing distributions of the generated narrative maps and the data set. Our analysis supports the overarching conclusion that the narrative maps extraction algorithm effectively encapsulates the framing distribution inherent in the broader data set. These findings suggest that the narrative maps, as extracted, have the potential to offer users valuable insights into the intricate framing dynamics embedded within the news narrative. 

However, when we examined whether the algorithm produces narrative maps with consistent framing, our analysis revealed inconsistencies within individual maps, despite capturing the broader distribution. For instance, one map shifted focus from public service issues to political and cultural issues across connected events. This indicates that the algorithm currently focuses on content similarities rather than coherent framing when extracting narratives.

Finally, we note that by explicitly evaluating framing in the context of computational narrative extraction and representation process, our study addresses the lack of studies regarding framing in the context of computational narrative extraction. The current model can indirectly capture framing information through cues in the underlying text representation. However, directly capturing and using framing information during the extraction process remains an open challenge. Future work should address this issue by exploring novel approaches and techniques that can explicitly incorporate framing information into the extraction pipeline, potentially improving the accuracy and contextual understanding of the extracted data.

\bibliography{sample-ceur}

\end{document}